\pdfoutput=1

\documentclass[11pt]{article}

\usepackage[table]{xcolor}
\usepackage{tcolorbox}

\usepackage[]{acl}

\usepackage{times}
\usepackage{latexsym}

\usepackage[T1]{fontenc}

\usepackage[utf8]{inputenc}

\usepackage{microtype}

\usepackage{inconsolata}

\usepackage{graphicx}

\usepackage{multirow}

\title{Correcting Hallucinations in News Summaries: Exploration of Self-Correcting LLM Methods with External Knowledge}

\author{Juraj Vladika, Ihsan Soydemir, Florian Matthes \\
  Technical University of Munich \\
  School of Computation, Information and Technology \\
  Department of Computer Science \\
  Garching, Germany \\
  \texttt{\{juraj.vladika, ihsan.soydemir, matthes\}@tum.de}}

\begin{document}
\maketitle
\begin{abstract}
While large language models (LLMs) have shown remarkable capabilities to generate coherent text, they suffer from the issue of hallucinations -- factually inaccurate statements. Among numerous approaches to tackle hallucinations, especially promising are the \textit{self-correcting} methods. They leverage the multi-turn nature of LLMs to iteratively generate verification questions inquiring additional evidence, answer them with internal or external knowledge, and use that to refine the original response with the new corrections. These methods have been explored for encyclopedic generation, but less so for domains like news summarization.
In this work, we investigate two state-of-the-art self-correcting systems by applying them to correct hallucinated summaries using evidence from three search engines.
We analyze the results and provide insights into systems' performance, revealing interesting practical findings on the benefits of search engine snippets and few-shot prompts, as well as high alignment of G-Eval and human evaluation.
\end{abstract}

\section{Introduction}
The advent of Large Language Models (LLMs) has revolutionized the field of Natural Language Processing (NLP), enabling models to perform complex tasks such as summarization and question answering with remarkable fluency \cite{wang2023pre}. 
While they can produce human-sounding text, LLMs are also prone to generating \textit{hallucinations} -- responses that sound convincing but are factually incorrect or misleading \cite{ji2023survey}. 
This limitation poses challenges for their reliability and adoption, especially in critical applications like law, healthcare, and news \cite{wang2023decodingtrust}. 

While numerous methods to counter hallucinations have been developed in recent years \cite{tonmoy2024comprehensivesurveyhallucinationmitigation}, many focus on pre-training and fine-tuning. For popular closed models like GPT or Gemini, the \textit{post-hoc correction} methods, which correct the initial response after it has been generated, are quite important. In particular, \textit{self-correcting} methods approach hallucination correction as a step-by-step process where the response is broken into smaller units and iteratively corrected using internal LLM knowledge or external sources \cite{kamoi-etal-2024-llms, vladika-etal-2025-step}. 

The effectiveness of these methods has been demonstrated for tasks like 
generating biographies or lists \cite{min-etal-2023-factscore, chern2023factoolfactualitydetectiongenerative}, but their application to news summarization remains underexplored. News articles are time-sensitive and factually dense, which underscores the need for correct summaries and effective fact-checking \cite{Graves2019FactCheckingAI, palic2019takelab}. 

 Furthermore, evidence retrieval is a crucial component of self-correcting systems -- many questions are open regarding which search engine to use, which snippets or article chunks to select, and how to best integrate them. Finally, the trade-off between balancing the faithfulness to original text with doing strong corrections is often neglected.

To explore these research gaps, we take two popular multi-step correction systems, CoVE \cite{dhuliawala-etal-2024-chain} and RARR \cite{gao-etal-2023-rarr}, augment them with external search engines, and apply them to correct hallucinated news summaries from the dataset SummEdits \cite{laban-etal-2023-summedits}. We compare the performance of different search engines and settings, three LLMs, two retrieval settings,  and the influence of prompts, uncovering important considerations for future.
We outline main challenges of these systems and provide future steps on how to improve them. Code and data is available in a GitHub repository.\footnote{\url{https://github.com/jvladika/HalluCorrect}}



\section{Related Work}
Hallucinations are a common problem in natural language generation (NLG) tasks, including abstractive text summarization \cite{ji2023survey, afzal2023challenges}. A survey by \citet{zhang2023sirens} divides hallucinations into input-conflicting, context-conflicting, and fact-conflicting. The focus of our work lies in fact-conflicting, which are hallucinations where facts in output contradict the world knowledge. 
While hallucinations can be observed by looking at the uncertainty in model's logits \cite{varshney2023stitch}, this is only possible for open-source models. In closed models such as ChatGPT, factuality has to be assessed through textual output. This has led to the rise of \textit{fact-checking} mechanisms \cite{vladika-matthes-2023-scientific, wang-etal-2024-factcheck, zhang2025knowhalu}, as well as \textit{self-correcting} LLM techniques \cite{kamoi-etal-2024-llms}.


The multi-step self-correcting LLM methods tend to base the corrections on internal LLM knowledge \cite{madaan2023selfrefine, kim2023language}. For external search, usually only Wikipedia \cite{gou2024critic} or Google search \cite{wei2024longform} is used. 
It is often applied to tasks like generating biographies. 
Abstractive summarization of news articles often contains factual errors \cite{tang2022understanding}. 
For news summaries, methods such as text infilling 
with smaller LMs 
 \cite{balachandran-etal-2022-correcting} 
 or entity linking to graphs \cite{dong-etal-2022-faithful} 
 have been explored to correct errors. 
Application of iterative self-correcting methods to the news domain is still mostly missing.

We augment the two self-correcting methods, CoVe and RARR, to support external search. Our study is among the first to explore this type of methods for news, to evaluate three different search engines, changes in snippets and full-text retrieval, and to compare closed with open base LLMs. 

\section{Systems}
In our study, we use two systems designed to detect and iteratively correct hallucinations, both of which have demonstrated strong results and gained popularity: Chain-of-Verification (CoVe) and Retrofit Attribution using Research and Revision (RARR). 

Both systems follow the same workflow: (1) Get Initial Response, (2) Generate Verification Questions (to help self-correct any errors), (3) Answer Questions (using evidence from internal knowledge or search engine), (4) Rewrite Response (with previous answers and any found inconsistencies).

Given the baseline response \textit{b}, there are \textit{k} generated follow-up questions $q_1, ..., q_k$, which try to gather more information related to the response \textit{b}. This is generated using a base LLM and a prompt $M_q$. Afterward, evidence $e$ for each question $q$ is retrieved from the source $s$ using the method $R(q,s)$, where \textit{s} can be internal LLM knowledge, gold news article, or external search engine. This collected evidence is used as input with questions to the answering model $M_a(q, e)$, which gives answers $a_1, ..., a_k$. Finally, baseline response and answers are given to the refinement model $M_r(b, a)$, which outputs the final refined response $r$. All prompts for \textit{M} are in Appendix \ref{sec:appendix}.

The difference between models is in prompts used to generate and answer the questions, and perform the final refinement. Also, CoVe is zero-shot, while RARR is based on few-shot examples.




\section{Setup and Experiments}
The LLM used in most experiments is \textit{GPT-4o-mini-2024-07-18} \cite{openai2024gpt4technicalreport}. It was queried through OpenAI API. Any encoder-only models were run on one Nvidia V100 GPU with 16GB VRAM for one computation hour.


\subsection{Dataset}
\paragraph{SummEdits} \cite{laban-etal-2023-summedits} is a benchmark dataset of hallucinated text summaries in many domains. The dataset was constructed by first perturbing named entities and relations in summaries and then passing to humans for annotation on whether the summaries are factual or not. We take the subset \textit{news} (constructed from top Google News 2023 articles), consisting of 819 summaries. While the original intent of benchmark was to evaluate hallucination detection ability of LLMs, we repurpose it for hallucination detection with fact correction.



\begin{table*}[htpb]
\small
	\centering
	\rowcolors{2}{gray!20}{white}
    
		
		\begin{tabular}{c|c|cc|ccc|ccc}
			
	\rowcolor{gray!40}
	\textbf{verification} & \textbf{evidence} & \multicolumn{2}{c}{\textbf{simple}} & \multicolumn{3}{c}{\textbf{NLI}} & 
	\multicolumn{3}{c}{\textbf{G-Eval $\uparrow$}}  \\
	
	\rowcolor{gray!40}
	\textbf{system} &  \textbf{source}  & \textbf{NED $\downarrow$}       & \textbf{Sem. $\uparrow$}      & \textbf{Ent.~$\uparrow$}       & \textbf{Neu.}       & \textbf{Con.~$\downarrow$}       & \textbf{Overall}       & \textbf{Factual.}       & \textbf{Relev.}     \\
	\hline
\textbf{CoVE} & GPT 4o mini & 0.51 & 81 &  30 & 28 & 42 & 50 & 45 & 49 \\ 
     \textbf{RARR} & GPT 4o mini  & 0.10 & 94 & 45 & 15 & 40  & 65 & 62 & 70 \\  \hline \hline
 
 \textbf{CoVE} &  Google (snip.)  & 0.51 & 84 &  \underline{41} & 25 & 34 & 56 & 50 & 59 \\  
  \textbf{CoVE} & Bing (snip.)  & 0.55 & 81 &  37 & 28 & 35 & 49 & 46 & 51 \\  
   \textbf{CoVE} & DDG (snip.)  & 0.54 & 80 &  31 & 28 & 41 & 47 & 42 & 47 \\  \hline
 \textbf{RARR} &  Google (full) &  0.33 & 91 & 24 & 46 & \textbf{30} & 64 & 51 & 68 \\
     \textbf{RARR} &  Bing (full)  & 0.32 & 92 & 28 & 40 & \underline{32}  & 63 & 50 & 68 \\ 
     \textbf{RARR} &  DDG (full)  & 0.34 & 91 & 27 & 41 & 32  & 64 & 50 & 68 \\ \hline
     \textbf{RARR} &  Google (snip.) & \underline{0.24} & \underline{93} & 40 & 28 & 32  & \underline{67} & \underline{56} & \underline{72} \\
     \textbf{RARR} &  Bing (snip.) & \textbf{0.14} & \textbf{95} & \textbf{49} & 16 & 35  & \textbf{69} & \textbf{60} & \textbf{73} \\ 
     \textbf{RARR} &  DDG (snip.)  & 0.25 & 92 & 32 & 28 & 40  & 60 & 49 & 62 \\ \hline  \hline
     \textbf{CoVE} & gold article  & 0.49 & 88 &  43 & 39 & 18 & 70 & 63 & 76 \\ 
     \textbf{RARR} & gold article  & 0.21 & 94 & 47 & 34 & 19  & 75 & 67 & 83 \\ \hline
     
    		\end{tabular}
	
	\caption{\label{tab:results_engines} Results of CoVE and RARR on SummEdits using three different search engines. NED refers to normalized edit distance, Sem. to average cosine semantic similarity, NLI scores to average prediction probability for entailment, neutral, and contradiction. 
 The best score for each metric is in \textbf{bold}, while the second best is \underline{underlined}.}
	
\end{table*}

\subsection{Evaluation Methods}

Since the dataset provides gold (human-written) summaries, we use them as reference answers.

We measure string dissimilarity using the Levenshtein normalized edit distance (\textbf{NED}) \cite{yujian2007normalized}. This metric is not ideal because even one word difference can be a major hallucination. Therefore, we compare the semantic similarity (\textbf{Sem.}) between the gold and output summary by embedding them with the model SimCSE \cite{gao-etal-2021-simcse} and calculating the cosine similarity.



\textbf{NLI Score} is a metric that utilizes the concept of natural language inference (NLI), or entailment recognition, by using the reference answer as the hypothesis and the generated answer as the premise. The intuition behind this approach is that a good answer should logically entail the reference answer. Using NLI this way has been done for evaluating the quality of summaries \cite{mishra-etal-2021-looking, laban-etal-2022-summac, steen-etal-2023-little}. Following this approach, we use the model DeBERTa-v3 \cite{he2023debertav}, 
We use the version fine-tuned on a wide array of NLI datasets, which works well for long text \cite{laurer2024less}.
This model predicts three scores (entailment, neutral, contradiction) and we report the average score across the whole dataset.
 
\textbf{G-Eval} \cite{liu-etal-2023-g} is a framework based on LLM prompting with chain-of-thoughts to evaluate the quality of generated texts in a form-filling paradigm. It is one of the most popular "LLM-as-judge" metrics \cite{zheng2023judging}, which evaluate the LLM output with an LLM using finely crafted LLM prompts (see Appendix \ref{sec:appendix}) and take the numerical output as final score. 
We evaluate three aspects: relevance, factuality, and overall quality.

\textbf{Human Evaluation.} We perform human evaluation with 25 participants. They were shown 10 gold summaries and refined summaries by RARR and CoVe, and rated for each the overall quality (based on factuality and relevance) from 1 to 10 and the entailment relation for each summary, amounting to 1000 ratings (see more details in Appendix~\ref{sec:human_eval}).


\subsection{Search Engines}
\textbf{Google} is the world's most widely used search engine. It offers the API service Google Programmable Search Engine, 
which queries the search engine and returns results as links and snippets. The price is 5 US dollars per 1,000 queries.  \textbf{Bing} is the flagship search engine from Microsoft.
We use it via Bing Web Search API provided by the Azure platform
for the price of 10 USD per 1,000 transactions. \textbf{DuckDuckGo} is a smaller search engine, aiming to help protect online privacy. While the coverage is lower than the other two engines, its usage through API is completely free.
We query it through the Python package \textit{duckduckgo-search}.\footnote{\url{https://pypi.org/project/duckduckgo-search/}} 

We use the search results of these search engines in two settings: chunks from \textit{full articles} and \textit{snippets}. All search engines return results for the query with links to articles included in top results. In the full-article setting, we parse the text from HTML of the article, split into chunks, embed with SimCSE, and use cosine similarity to the query to select top 5 passages. We concatenate these passages and use them as input evidence. In addition to links, all search engines provide snippets that answer the query and highlight the most important part from the respective article. We use the top 5 snippets and concatenate them, using them as input evidence.

\begin{table*}[htpb]
\centering
\small
\begin{tabular}{p{30mm}p{115mm}}
\hline
\textbf{Gold Summary} &  The James Webb Space Telescope captured a new image of Pandora's Cluster, a megacluster of galaxies that allowed astronomers to peer into the distant universe and observe never-before-seen details."  \\[0.7ex]

\textbf{Hallucinated \newline Summary \newline (input to system)}  & The James Webb Space Telescope captured a \textbf{old} image of Pandora's Cluster, a megacluster of galaxies that allowed \textbf{biologists} to peer into the distant universe and observe never-before-seen details. \\

\hline

CoVe with \textbf{Bing} snip. & The James Webb Space Telescope is primarily designed for astronomical observations, \textbf{not biological observations}. It is used to study the formation of stars, galaxies, and planetary systems, as well as to investigate the origins of life and the universe. It cannot be used for biological observations. The telescope studies galaxy formation and evolution, star formation and planet formation, planetary systems, and the origins of life. \textbf{It did capture an image of Pandora's Cluster.}
  \\[0.7ex]

CoVe with \textbf{Google} snip. & The James Webb Space Telescope is primarily designed for astronomical observations, specifically to study the formation of stars and galaxies, the evolution of galaxies, and the formation of planetary systems. \textbf{It is not designed for biological observations.} Therefore, it does not allow biologists to observe never-before-seen details in the distant universe. However, it is \textbf{\textit{capable of capturing}} images of Pandora's Cluster. \\[0.7ex]

 CoVe with \textbf{DDG} snip. & The James Webb Space Telescope is primarily designed for astronomical observations, \textbf{not biological observations}. It is used to study the formation of stars, galaxies, and planetary systems, among other astrophysical phenomena. It can observe details in the distant universe and has captured \textit{images of megaclusters of galaxies.}
  \\   \hline

\end{tabular}
\caption{Example of final refined responses from CoVe using the search snippets from three different search engines. All results correctly identified the error with biologists, although only Bing properly reported on the image of Pandora's Cluster being captured.}
\label{tab:examples_engines}
\end{table*}

\section{Results and Discussion}
Table \ref{tab:results_engines} shows the average results of all metrics for the two systems on SummEdits. Qualitative insights are found in Tables \ref{tab:examples_engines} and \ref{tab:examples_systems}.

\paragraph{Internal vs.~External Knowledge.} The first two rows of Tab.~\ref{tab:results_engines} used internal LLM knowledge to answer verification question. While this led to moderate performance, results with search engines were higher for both systems -- showing the \textbf{need for external search} for effective factual error correction. 

The last two rows show the baseline of using the original (gold) news article as input evidence. It had the highest G-Eval scores, highlighting the key role of precise evidence for effective corrections.

\paragraph{Choice of Search Engine.} As seen in Table \ref{tab:results_engines}, Google snippets performed the best for CoVE but Bing outperformed it on RARR for the full-article setting. The highest performance overall was achieved by Bing snippets with RARR, as measured by six different metrics. This shows the promising potential of Bing, which is underexplored in existing studies. DuckDuckGo also achieved decent but lower performance. 

Table \ref{tab:examples_engines} shows examples of refined responses from CoVe 
with the three engines. All three engines successfully identified the hallucination involving biologists. Still, they found different extra information that was included in the refined response, with Bing the only one pinpointing the correct telescope discovery. This shows the engines get similar results but Bing usually led to best corrections overall, because for generated queries Bing provided longer and more informative snippets. 

Still, the high price of Google and Bing APIs could be a hurdle for large-scale usage. DuckDuckGo achieved adequate performance for \$0 and, thus, offers a promising free alternative. Future work could explore additional search filters or filtering of results by trustworthy domains.

\paragraph{Snippets or Full-Article Chunks?} When looking at RARR scores of NED, sem.~similarity, and G-Eval for snippets and full articles, all are slightly better for the snippets setting. Still, the average NLI scores reveal the full picture -- the setting with full articles had high NLI neutral scores. This is because the responses were often refined with irrelevant evidence, whereas the snippet setting produced responses that had a higher NLI-entailment score. The contradiction scores were similar in both. This shows that \textbf{snippets} are usually more \textbf{on point and related to the actual search query}, while using the \textbf{full articles} can lead to selecting \textbf{noisy or irrelevant passages from articles}.

Another interesting finding is the general similarity of overall scores, including NLI scores, for the setting with full articles -- this shows that all three search engines mostly found the same articles among its top results and then the similarity function selected same passages from those articles.

\begin{table*}[htpb]
\centering
\small
\begin{tabular}{p{33mm}p{112mm}}
\hline
\textbf{Gold Summary} &  The founder of the troubled digital start-up Ozy Media has been arrested on fraud charges as part of a scheme to prop up the financially struggling company.  \\[0.8ex]

\textbf{Hallucinated Summary (input to system)}  & The founder of the \textbf{triumphant} digital start-up Ozy Media has been arrested for fraud charges as part of a scheme to \textbf{sell} the financially \textbf{thriving} company Ozy Media.  \\ [0.8ex]

\hline
\textbf{RARR} (Bing snip.) & 
The founder of the \textbf{ailing} digital start-up Ozy Media has been arrested on fraud charges as part of a plan to \textbf{support} the financially \textbf{struggling} company.
\\[0.8ex]

\textbf{CoVe} (Bing snip.) & 
The CEO of Ozy Media, Carlos Watson, has been arrested on fraud charges. Ozy Media is facing financial \textbf{struggles} and there is a scheme in place to \textbf{prop up} the company. However, there is no information or news updates regarding the CEO's arrest. The current status of Ozy Media is that it has shut down, and its CEO has been accused of deceiving potential investors. The company's chairman, Marc Lasry, has stepped down.
  \\   \hline

\end{tabular}
\caption{Example of an instance from SummEdits with final outputs of CoVE and RARR, both based on Bing search snippets. The RARR correction was a lot more concise, while CoVe was more informative and detailed.}
\label{tab:examples_systems}
\end{table*}

\paragraph{Zero-shot or Few-shot?} Table \ref{tab:examples_systems} shows an instance from SummEdits with the gold, hallucinated, and refined summaries by both systems. Both used Bing as the search engine and, thus, both managed to correct factual errors in the input summary (e.g., \textit{struggling} instead of \textit{thriving}). Still, it is evident that RARR returned a summary close in form to the input summary, whereas CoVe augmented the summary with additional information found on Bing. 

This difference in length is the consequence of the fact that RARR uses six examples in its few-shot prompt, while CoVe does not use any examples.
CoVe also sometimes returned summaries similar to the input summary with minimal changes, however it often returned a lot longer summaries. Long summaries do not necessarily imply hallucinations, but can be summaries with additional context for readers. This points to the fact that \textbf{few-shot} prompts are better if the end goal is to \textbf{preserve the faithfulness to the original draft}, while \textbf{zero-shot} relaxed prompts are better when \textbf{adding additional context and making bold edits is preferred}. The few-shot examples are general-domain, so the findings are not just for news.

\paragraph{Open LLMs.} We also ran experiments with LLaMa 3.1 (70B), results are in Table~\ref{tab:open_llms}. For RARR, it had on average weaker scores than GPT 4o-mini, but came quite close, confirming the recent trend of open models closing the gap to closed competitors. For CoVe, which does not have few-shot examples, it generated a lot longer final refined responses than GPT, with lots of detailed explanations. This led to increased G-Eval (Overall \& Relevance) and NLI metrics, since these metrics favor information-heavy summaries, but the G-Eval factuality score heavily decreased and summaries were too complex.  We additionally ran Mixtral 8x7B with internal knowledge, but it underperformed compared to both Llama and GPT.   Future work could explore more open LLMs and evaluate user-centric text quality aspects like readability.

\subsection{Human Evaluation}
The mean human scores for quality of 10 examples with Bing snippets for 25 participants were \textbf{0.68} for RARR and \textbf{0.54} for CoVe, showing \textbf{users preferred RARR} refinements. The mean G-Eval score for these 10 examples were \textbf{0.65} and \textbf{0.52}, respectively. This shows an impressively \textbf{high alignment of humans with G-Eval} (Pearson correlation coefficient 0.87, p<1\%), with the average difference of 3\%.  Our custom prompts for factuality and relevancy have a high potential for future use, and this positions G-Eval as a promising metric to use when human annotations are not available due to time and costs. For NLI, the alignment was decent but less apparent -- DeBERTa overpredicted the neutral class, while human annotators favored the entailment class. Ideally, the DeBERTa-NLI model should be fine-tuned on examples focusing on hallucination detection. More details on human evaluation are in Appendix~\ref{sec:human_eval}.

\section{Conclusion and Future Work}
In this study, we explored the impact of different evidence sources and search engines on the performance of two SotA systems for post-hoc hallucination correction, CoVe and RARR, for news summaries. Our detailed results show that zero-shot correction systems like CoVe yield more expressive and bold corrections that change the style, while few-shot systems like RARR optimize for faithfulness to the original text and this was favored by humans in evaluation. Additionally, G-Eval metric was highly aligned with humans.
We also found that Bing's search snippets led to most informative corrections, followed closely by Google, but DuckDuckGo can be a viable alternative due to its free API and decent performance. 
We envision future work focusing on enhancing retrieval with structured queries and assessing evidence reliability.

\section*{Limitations}
An important limitation lies in the fact that all modules of the iterative self-correcting systems rely on using LLMs, which comes with its own set of challenges. The generated follow-up questions are not always perfect or precise, the generated answers from snippets can be off-point, and the final refinement of responses can be too excessive. Future work could explore how to incorporate more controllable generation or structured and rule-based techniques for correcting the output.

Another limitation comes from the high complexity of the system and reliance on calls to external APIs, including LLM APIs and search engine APIs. This can inevitably lead to slow processing speed of these systems when compared to approaches that use smaller encoder-only models or rule-based techniques. Still, we were forced to rely on API calls to LLMs due to our hardware resource limitations.
Other lines of work could explore how to better incorporate open and local models into the workflow, for better accountability and faster processing time.

Finally, our work deals only with the news domain, which could limit the generalizability of findings to other domains and use cases.

\bibliography{custom}

\appendix

\section{Human Evaluation}
\label{sec:human_eval}
The main goal of the human evaluation was to judge two automated metrics, NLI predictions and LLM-as-a-judge (G-Eval), by observing the alignment between human preference and machine evaluation results. All the evaluation responses and results are attached to the ARR submission.

\subsection{Study Format and Instructions}

User study was conducted with 25 participants. All participants are pursuing a master's degree or a PhD degree in computer science at authors' university. They were not monetarily compensated since they are in-house annotators from our school's department of computer science. All responses were anonymous and collected only for the purpose of this research study. Users were provided with instructions described in Table \ref{tab:instructions}.

\begin{figure*}[htpb]
  \centering
  {\includegraphics[width=0.80\textwidth]{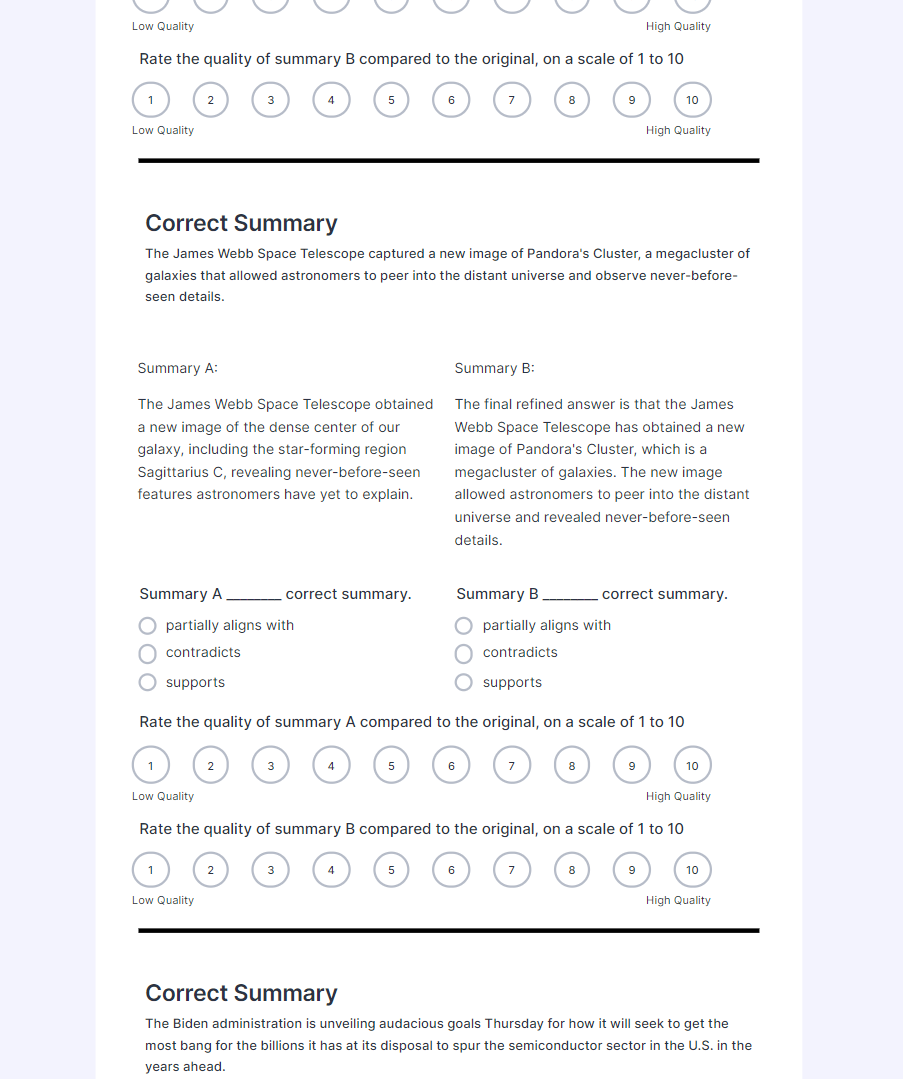}}
  \caption{A screenshot from Human Evaluation Form.}
  \label{fig:humaneval}
\end{figure*}

The survey was hosted as a questionnaire on the JotForm platform.\footnote{\url{https://www.jotform.com}} In total, there were 10 examples, where each example consisted of a correct summary, a hallucinated summary, a summary corrected by CoVe, a summary corrected by RARR, and 4 questions to answer. In Figure \ref{fig:humaneval}, a sample screenshot from the evaluation form is provided.

Users were asked to evaluate each of the two generated summaries in two aspects: overall quality and NLI relation. The overall quality was estimated by rating from 1 to 10 and it refers to (a) how factually accurate was the summary, and (b) how relevant and on-topic was it. The NLI (entailment) relation were mapped to NLI classes by asking the users whether the generated summary supports the gold summary (entailment), contradicts the gold summary (contradiction), or partially aligns with the gold summary (neutral).

In each example, we include samples from RARR or COVE as either summary A or B. Correct summary represents the ground truth summary from the SummEdits dataset. Summary A or B from self-correcting systems were generated using snippets from the Bing search engine. Both self-correcting systems were provided with the same hallucinated version of the correct summary and the pipeline for rewriting was ran.

\subsection{Overall Quality Results}

In the survey, the "overall quality" score was rated from 1 to 10 and it referred to how factual the summary was and how relevant (on-topic) it was, when compared to the original (gold summary). To evaluate the alignment between the G-Eval scores and human evaluations for the RARR and COVE methods, we analyzed the mean scores and their differences. Human scores are an average of 250 scores, normalized to the percentage value. Results are summarized in Table~\ref{table:humaneval}.

For RARR, average human score is 0.68, and average G-Eval score is 0.65. For the COVE method, average human score is 0.54, and average G-Eval score is 0.52.  G-Eval scores are slightly lower than human evaluations. These minor differences for both RARR and COVE suggest that G-Eval scores closely reflect human evaluations for both methods, with a deviation of ±3\%.

\begin{table}[htpb]
\hrulefill \\
\small \underline{Instructions} \\
Read the correct summary first. \\
Compare the correct summary with the generated Summary A and Summary B. \\
There are no right or wrong answers. \\
Both summaries can be good or bad. \\
For each summary (A and B), there are two types of questions:

    1. Choose the option that best fits the blank:
    \begin{itemize}
        \item \textbf{Contradicts}: Disagrees with the correct summary
        \item \textbf{Supports}: Agrees with the correct summary
        \item \textbf{Partially aligns with}: Only somewhat related or unrelated
    \end{itemize}
    2. Rate the Overall Quality (Factual accuracy + Relevance):
    \begin{itemize}
        \item \textbf{Factual accuracy}: Is it based on facts? Avoids misinformation?
        \item \textbf{Relevance}: Does the summary cover the main points? Not off-topic?
    \end{itemize}
\hrulefill 
\caption{Instructions that human annotators received.}
\label{tab:instructions}

\end{table}

\subsection{Natural Language Inference Results}

We also compared human evaluation and ground truth values for Natural Language Inference (NLI) across three categories: Entailment, Neutral, and Contradiction. As discussed before, DeBERTa-v3 model \cite{laurer2024less} is used for NLI evaluation. The results are presented in Table~\ref{table:nli_human}.

\begin{table*}[htpb]
\centering
\begin{tabular}{lccc}
\hline
\textbf{Method} & \textbf{Human Mean Score} & \textbf{G-Eval Score} & \textbf{Diff} \\
\hline
RARR & 0.68 & 0.65 & 0.03 \\
COVE & 0.54 & 0.52 & 0.02 \\
\hline
\end{tabular}
\caption{Alignment between G-Eval scores and human evaluations.}
\label{table:humaneval}
\end{table*}

\begin{table*}[htpb]
\centering
\begin{tabular}{lccc|ccc}
\hline
\textbf{Method} & \multicolumn{3}{c|}{\textbf{Human}} & \multicolumn{3}{c}{\textbf{NLI Model}} \\
\cline{2-7}
 & \textbf{Entailment} & \textbf{Neutral} & \textbf{Contradiction} & \textbf{Entailment} & \textbf{Neutral} & \textbf{Contradiction} \\
\hline
RARR & 45 & 40 & 15 & 30 & 49 & 21 \\
COVE & 31 & 37 & 32 & 28 & 47 & 25 \\
\hline
\end{tabular}
\caption{Comparison of Human Evaluation and NLI predictions}
\label{table:nli_human}
\end{table*}

For both self-correcting systems, there is a higher percentage of Entailment in human evaluations compared to the NLI model, particularly in RARR. Also, percentage of Neutral instances is lower in human evaluations. NLI model is more likely to classify instances as Neutral than humans. Contradiction shows higher percentages in human evaluations for COVE compared to the NLI model. Overall, as demonstrated by evaluation of experiments and human evaluation, RARR performs better than COVE in SummEdits dataset.

\subsection{Alignment between Automated Metrics and Human Scores}
Analyses indicate a strong alignment between G-Eval scores and human evaluations for both RARR and COVE methods in rating the overall quality aspect. This consistency means that G-Eval is a reliable tool for approximating human assessments. It can be used in scenarios where human evaluations are impractical when there are time or resource constraints.

When it comes to NLI, humans had a somewhat different feeling of which class to assign 
than the automated method. Differences between human evaluation and automated predictions were more evident than in case of G-Eval, although there was still an alignment in terms of predominant classes. This shows that while NLI is a decent metric, there is still room for improvement, possibly in terms of additionally fine-tuning the predictor model (DeBERTa) on further NLI datasets or datasets centered around the specific tasks of factuality and generation-quality prediction. Another option is using more complex models like LLMs for prediction, although they have been found to favor the entailment class as opposed to the neutral class in NLI predictions \cite{zhou-etal-2024-constructions}.

\section{Results with Open LLMs}
We additionally performed experiments with two popular open-source LLMs, Llama 3.1 (70B) \cite{grattafiori2024llama3herdmodels} and Mixtral 8x7B \cite{jiang2024mixtralexperts}, to test how well do they fare compared to GPT. The results are shown in Table \ref{tab:open_llms}. The models were prompted using the API endpoint of Together AI,\footnote{\url{https://www.together.ai/}} a platform that host popular open-source LLMs. All the settings we applied were the same as for GPT and Open AI's API, including temperature set to 0 for better reproducibility.

\begin{table*}[htpb]

	\centering
    
    \resizebox{\textwidth}{!}{%
		
		\begin{tabular}{c|c|c|cc|ccc|ccc}
			
	\rowcolor{gray!40}
	\textbf{Base} & \textbf{verification} & \textbf{evidence} & \multicolumn{2}{c}{\textbf{simple}} & \multicolumn{3}{c}{\textbf{NLI}} & 
	\multicolumn{3}{c}{\textbf{G-Eval}}  \\
	
	\rowcolor{gray!40}
	\textbf{LLM} & \textbf{system} &  \textbf{source}  & \textbf{NED}       & \textbf{Sem.}      & \textbf{Ent.}       & \textbf{Neu.}       & \textbf{Con.}       & \textbf{Overall}       & \textbf{Factual.}       & \textbf{Relev.}     \\
	\hline
     \multirow{2}{*}{Mixtral 8x7B} & \textbf{CoVE} & Mixtral & 0.77 & 74 &  30 & 48 & 22 & 64 & 42 & 59 \\ 
     & \textbf{RARR} & Mixtral  & 0.43 & 84 & 26 & 32 & 42  & 55 & 43 & 50 \\  \hline \hline
     \multirow{2}{*}{LLaMa 3.1 (70B)} & \textbf{CoVE} & Llama & 0.78 & 70 &  38 & 51 & \textbf{11} & 67 & 50 & 73 \\ 
     & \textbf{RARR} & Llama  & 0.20 & 94 & 39 & 24 & 37  & 63 & 59 & 71 \\  \hline \hline 
     \multirow{3}{*}{LLaMa 3.1 (70B)} & \textbf{} & Google & 0.78 & 75 &  43 & 44 & \underline{13} & \underline{67} & 47 & \underline{73} \\ 
     & \textbf{CoVE} & Bing  & 0.79 & 75 & 41 & 44 & 15  & \textbf{68} & 46 & \textbf{74} \\
      & \textbf{} & DDG  & 0.80 & 73 & 34 & 46 & 20  & 59 & 39 & 66 \\ \hline
     \multirow{3}{*}{LLaMa 3.1 (70B)} & \textbf{} & Google & 0.28 & 90 &  \textbf{46} & 24 & 30 & 66 & \textbf{62} & 72 \\ 
      & \textbf{RARR} & Bing & 0.33 & 88 &  \underline{44} & 28 & 28 & 64 & \underline{59} & 70 \\
       & \textbf{} & DDG & 0.42 & 84 &  34 & 26 & 40 & 54 & 48 & 58 \\\hline \hline
     
 \hline
     
    		\end{tabular}
	}
	
	\caption{\label{tab:open_llms} Results of CoVE and RARR on SummEdits using two open-source LLMs, Llama 3.1 and Mixtral. NED refers to normalized edit distance, Sem. to average cosine semantic similarity, NLI scores to average prediction probability for entailment, neutral, and contradiction. 
 The best score for each metric is in \textbf{bold}, while the second best is \underline{underlined}.}
	
\end{table*}

\section{Prompts}
\label{sec:appendix}

This appendix section 
provides the prompts used in the CoVe system in Table \ref{tab:cove_prompts} and for the RARR system in Tables \ref{tab:rarr_prompts_part1} and \ref{tab:rarr_prompts_part2}. Additionally, the prompts used for the LLM-as-judge metric G-Eval are given in Table \ref{tab:geval}.

\begin{table*}[htpb]
\footnotesize

\centering
\begin{tabular}{p{28mm}p{122mm}}
\hline
\textbf{Use Case} & \textbf{Prompt Content}\\
\hline
Generate verification question (template) & Your task is to create a verification question based on the below question provided. \newline
Example Question: Who are some movie actors who were born in Boston?\newline
Example Verification Question: Was [movie actor] born in [Boston]\newline
Explanation: In the above example the verification question focused only on the ANSWER\_ENTITY (name of the movie actor) and QUESTION\_ENTITY (birth place).\newline
Similarly you need to focus on the ANSWER\_ENTITY and QUESTION\_ENTITY from the actual question and generate verification question.\newline

Actual Question: {original\_question}\newline

Final Verification Question:
\\
\hline

Generate verification question & Your task is to create verification questions based on the below original question and the baseline response. The verification questions are meant for verifying the factual accuracy in the baseline response. Output should be numbered list of verification questions. \newline

Actual Question: {original\_question}\newline
Baseline Response: {baseline\_response}

Final Verification Questions:
    \\ \hline

Answer verification question & Answer the following question correctly based on the provided context. The question could be tricky as well, so think step by step and answer it correctly. \newline

Context: {search\_result}

Question: {verification\_question}

Answer:
\\ \hline

Refine the original response & Given the below `Original Query` and `Baseline Answer`, analyze the `Verification Questions \& Answers` to finally filter the refined answer.\newline

Original Query: {original\_question}

Baseline Answer: {baseline\_response}

Verification Questions \& Answer Pairs:
{verification\_answers}\newline

Final Refined Answer:
\\ \hline

\end{tabular}
\caption{Overview of prompts used for the Chain-of-Verification (CoVE) system.}
\label{tab:cove_prompts}
\end{table*}

\begin{table*}[htpb]
\footnotesize

\centering
\begin{tabular}{p{28mm}p{122mm}}
\hline
\textbf{Use Case} & \textbf{Prompt Content}\\
\hline

Generate verification question & I will check things you said and ask questions.\newline
You said: Your nose switches back and forth between nostrils. When you sleep, you switch about every 45 minutes. This is to prevent a buildup of mucus. It’s called the nasal cycle.\newline

To verify it,\newline
1. I googled: Does your nose switch between nostrils?\newline
2. I googled: How often does your nostrils switch?\newline
3. I googled: Why does your nostril switch?\newline
4. I googled: What is nasal cycle?\newline

You said: The Stanford Prison Experiment was conducted in the basement of Encina Hall, Stanford’s psychology building.\newline
To verify it,\newline
1. I googled: Where was Stanford Prison Experiment was conducted?\newline
\textit{(four more examples)}

You said: {claim}\newline
To verify it,
\\ \hline

Answer verification question & I will check some things you said.\newline

1. You said: Your nose switches back and forth between nostrils. When you sleep, you switch about every 45 minutes. This is to prevent a buildup of mucus. It’s called the nasal cycle.\newline
2. I checked: How often do your nostrils switch?\newline
3. I found this article: Although we don’t usually notice it, during the nasal cycle one nostril becomes congested and thus contributes less to airflow, while the other becomes decongested. On average, the congestion pattern switches about every 2 hours, according to a small 2016 study published in the journal PLOS One.\newline
4. Reasoning: The article said the nose’s switching time is about every 2 hours, and you said the nose's switching time is about every 45 minutes.\newline
5. Therefore: This disagrees with what you said.\newline

1. You said: The Little House books were written by Laura Ingalls Wilder. The books were published by HarperCollins.\newline
2. I checked: Who published the Little House books?\newline
3. I found this article: These are the books that started it all -- the stories that captured the hearts and imaginations of children and young adults worldwide. Written by Laura Ingalls Wilder and published by HarperCollins, these beloved books remain a favorite to this day.\newline
4. Reasoning: The article said the Little House books were published by HarperCollins and you said the books were published by HarperCollins.\newline
5. Therefore: This agrees with what you said.\newline
\textit{(four more examples)}

1. You said: {claim}\newline
2. I checked: {query}\newline
3. I found this article: {evidence}\newline
4. Reasoning:\newline
\\ \hline

\end{tabular}
\caption{Overview of prompts for verification question generation and answering used for the RARR system.}
\label{tab:rarr_prompts_part1}
\end{table*}

\begin{table*}[h]
\footnotesize

\centering
\begin{tabular}{p{28mm}p{122mm}}
\hline
\textbf{Use Case} & \textbf{Prompt Content}\\
\hline

Refine the original response & I will fix some things you said.\newline

1. You said: Your nose switches back and forth between nostrils. When you sleep, you switch about every 45 minutes. This is to prevent a buildup of mucus. It’s called the nasal cycle.\newline
2. I checked: How often do your nostrils switch?\newline
3. I found this article: Although we don’t usually notice it, during the nasal cycle one nostril becomes congested and thus contributes less to airflow, while the other becomes decongested. On average, the congestion pattern switches about every 2 hours, according to a small 2016 study published in the journal PLOS One.\newline
4. This suggests 45 minutes switch time in your statement is wrong.\newline
5. My fix: Your nose switches back and forth between nostrils. When you sleep, you switch about every 2 hours. This is to prevent a buildup of mucus. It’s called the nasal cycle.\newline

1. You said: In the battles of Lexington and Concord, the British side was led by General Thomas Hall.\newline
2. I checked: Who led the British side in the battle of Lexington and Concord?\newline
3. I found this article: Interesting Facts about the Battles of Lexington and Concord. The British were led by Lieutenant Colonel Francis Smith. There were 700 British regulars.\newline
4. This suggests General Thomas Hall in your statement is wrong.\newline
5. My fix: In the battles of Lexington and Concord, the British side was led by Lieutenant Colonel Francis Smith.\newline

\textit{(four more examples)}\newline

1. You said: {claim}\newline
2. I checked: {query}\newline
3. I found this article: {evidence}\newline
4. This suggests
\\ \hline

\end{tabular}
\caption{Overview of prompts for response refinement used for the RARR system.}
\label{tab:rarr_prompts_part2}
\end{table*}

\begin{table*}[htpb]
\footnotesize

\centering
\begin{tabular}{p{28mm}p{122mm}}
\hline
\textbf{Evaluated Aspect} & \textbf{Prompt Content}\\
\hline

Factuality & Evaluate if the actual output contains hallucinated information not present in the input.\newline

STEPS: Identify any claims or statements in the 'actual output'.\newline
Compare each claim with the 'input' to check for the presence of supporting information.\newline
Mark any claims that are not supported by the 'input' as hallucinated.\newline
Penalize heavily for any introduction of new, unsupported facts.

\\ \hline

Relevance & 
Evaluate the relevancy of the actual output to the input.\newline

STEPS: Check if 'actual output' directly addresses the query or topic presented in 'input'.\newline
Penalize responses that are off-topic or provide irrelevant information.

\\ \hline
Overall & Evaluate the overall quality and correctness of the actual output compared to the input.\newline

STEPS: Assess if the 'actual output' provides a coherent and accurate response to 'input'.\newline
Penalize factual inaccuracies, grammatical errors, and unclear language.

\\ \hline

\end{tabular}
\caption{Overview of prompts used for the G-Eval metric.}
\label{tab:geval}
\end{table*}

\end{document}